%% file: ICLR19.tex
\documentclass{article}
\usepackage{iclr2019_conference,times}

\input{math_commands.tex}

\usepackage{hyperref}
\usepackage{url}
\usepackage{color, colortbl}
\usepackage{longtable}
\usepackage{breakcites}
\usepackage{pdfpages}
\usepackage{rotating}
\usepackage{tikz} 
\usetikzlibrary{arrows,calc,fit, patterns}
\definecolor{blue}{rgb}{0.5,0.5,0.5}

\usetikzlibrary{arrows.meta}

\usepackage[english]{babel}
\usepackage[T1]{fontenc}
\usepackage[utf8]{inputenc} 
\usepackage{textcomp}

\usepackage{stmaryrd} 

\usepackage[a4paper,top=3cm,bottom=2cm,left=3cm,right=3cm,marginparwidth=1.75cm]{geometry}

\usepackage{amsmath}
\usepackage{amsfonts}
\usepackage{adjustbox}
\usepackage{graphicx}
\usepackage{wrapfig, lipsum}
\graphicspath{{images/}{../images/}}
\usepackage{subfiles}
\usepackage{blindtext}
\usepackage{subcaption}
\usepackage{mathtools}
\usepackage{listings}
\usepackage{xspace}

\usepackage{longtable} 
\usepackage{rotating} 
\usepackage{dsfont} 

\definecolor{babyblueeyes}{rgb}{0.63, 0.79, 0.95}
\definecolor{blanchedalmond}{rgb}{1.0, 0.92, 0.8}
\definecolor{tangerine}{rgb}{0.95, 0.52, 0.0}

\usepackage[colorinlistoftodos]{todonotes} 

\newcommand{\gtcorr}{\textit{GTC}\xspace}
\newcommand{\gtcorrmean}{$GTC_{mean}$\xspace}

\title{Decoupling feature extraction from policy learning: assessing benefits of state representation learning in goal based robotics}

\author{Antonin Raffin, Ashley Hill, Ren\'e Traor\'{e}, Timoth{\'{e}}e Lesort, Natalia D{\'{\i}}az-Rodr{\'{\i}}guez, David Filliat  \\
U2IS \& INRIA FLOWERS Team\thanks{http://flowers.inria.fr}, \\ 
 ENSTA ParisTech, Palaiseau, France \\
\texttt{\{antonin.raffin, ashley.hill, rene.traore, timothee.lesort, natalia.diaz,} \\
\texttt{david.filliat\}@ensta-paristech.fr} \\
}

\iclrfinalcopy 
\begin{document}
\maketitle

\begin{abstract}

Scaling end-to-end reinforcement learning to control real robots from vision presents a series of challenges, in particular in terms of sample efficiency. Against end-to-end learning, state representation learning can help learn a compact, efficient and relevant representation of states that speeds up policy learning, reducing the number of samples needed, and that is easier to interpret. We evaluate several state representation learning methods on goal based robotics tasks and propose a new unsupervised model that stacks representations and combines strengths of several of these approaches. This method encodes all the relevant features, performs on par or better than end-to-end learning with better sample efficiency, and is robust to hyper-parameters change.

\end{abstract}



 
\section{Introduction}
\subfile{sections/introduction}

\subfile{sections/related-work}
\subfile{sections/specifications}
\subfile{sections/model-building}

\subfile{sections/experiments}

\subfile{sections/conclusions}



\clearpage

\bibliography{references}
\bibliographystyle{iclr2019_conference}

\subfile{sections/appendix}

\end{document}

%% file: math_commands.tex
\usepackage{amsmath,amsfonts,bm}

\def\eqref#1{equation~\ref{#1}}

\def\1{\bm{1}}

\DeclareMathAlphabet{\mathsfit}{\encodingdefault}{\sfdefault}{m}{sl}
\SetMathAlphabet{\mathsfit}{bold}{\encodingdefault}{\sfdefault}{bx}{n}



%% file: sections/introduction.tex

A common strategy to learn controllers in robotics is to design a reward function that defines the task and search for a policy that maximizes the collected rewards with a Reinforcement Learning (RL) approach. 

In RL, the controlled system (environment and robot) is defined by a state $s_t$, i.e., the relevant variables for a controller, often of low dimension (e.g., positions of a robot and a target). At a given state $s_t$, the agent will receive an \textit{observation} $o_t$ from the environment and a reward $r_t$. In some applications, the observation may be directly the state, but in the general case, the observation is raw sensor data (e.g., images from the robot camera). RL must then learn a policy that takes observations as input and returns the action $a_t$ that maximizes expected return.

When the state is not directly accessible, RL should recover it from the observation to learn a good control policy. 
State representation learning (SRL) \citep{Lesort18} aims at learning to extract those states separately from learning the RL policy. 

In this paper, we demonstrate the utility of SRL in goal-based robotics tasks, i.e. the controlled agent is a robot, the reward is sparse and only depends on the previous state and taken action, not on a succession of states (therefore excluding tasks like walking or running). We investigate the benefit of different ways of combining state of the art SRL approaches on policy learning for various goal based robotics tasks. 




To summarize our contribution, in this paper,
 we show the usefulness of decoupling feature extraction from policy learning 
and that random features provide a good baseline. We propose a new way of combining approaches by stacking state representations instead of mixing them, that allows to mitigate the problem of conflicting objectives and favors representation disentanglement. 
Finally, we investigate the influence of hyper-parameters when learning a state representation (Section~\ref{sec:ablation-study}). 


%% file: sections/related-work.tex
\section{Related Work}
\label{sec:related-work}


In reinforcement learning, a classic preparatory approach is to design some features by hand, in order to facilitate policy learning. However the manual design may be difficult, laborious and requires domain knowledge. Hence, this process can be automated using methods that are able to learn these features (also called representations) \citep{Bohmer15, Singh12, Boots2011}. This problem is commonly called \textit{State Representation Learning} (SRL).  
We can define it more precisely as a particular kind of representation learning where the learned features are in low dimension, evolve through time, and are influenced by the actions of an agent \citep{Lesort18}.

SRL is used as a preliminary step for learning a control policy. The representation is learned on data gathered in the environment by an exploration policy. 
One particular advantage of this step is that it reduces the search space and gives to reinforcement learning an informative representation, instead of raw data (e.g. pixels). This allows to solve tasks more efficiently~\citep{lange2012autonomous, wahlstrom2015pixels, Munk16}. We refer to \citep{Lesort18} for a complete review of SRL for control.

Since in robotics, experiments are time consuming and the availability of robots is limited by cost and maintenance constraints, several approaches prefer to iterate at first in simulation to learn robotics tasks~\citep{Jonschkowski15, Watter15, Curran16, Lesort19, Jonschkowski17}. Our proposal is based along this line. Our setting is similar to the one used in Hindsight Experience Replay~\citep{Andrychowicz2017} that tackles the problem of solving goal-based robotics tasks with sparse reward. However, in their experiments, the agent has a direct access to the positions of the controlled robot and target when in our work, we dont have access to direct position and use raw pixels as input. The extraction of relevant positions must be learned by a SRL method. 

%% file: sections/model-building.tex
\section{Building a method to encode the robot and target positions}
\label{sec:model-building}


Given the general objectives defined in the previous section, we now propose a way to combine several approaches by tackling one objective at a time, using a particular context for a concrete illustration. This part aims at giving insights on the different SRL methods, taking advantage of goal-based robotics tasks as an application example.

One important aspect to encode for RL is the state of the controlled agent. In the context of goal-based robotics tasks, it corresponds to the robot position.
A simple method consists of using an \textit{inverse dynamics objective}: given the current $s_t$ and next state $s_{t+1}$, the task is to predict the taken action $a_t$. The type of dynamics learned is constrained by the network architecture. For instance, using a linear model imposes linear dynamics.
The state representation learned encodes only controllable elements of the environment. Here, the robot is part of them. However, the features extracted by an inverse model are not always \textit{sufficient}: in our case, they do not encode the position of the target since the agent cannot act on it.

Since learning to extract the robot position is not enough to solve goal-based tasks, we need to add extra objective functions in order to encode the position of the target object. In this section, we consider two of them: minimizing a reconstruction error (auto-encoder model) or a reward prediction loss.
\begin{itemize}
\item \textit{Auto-encoder:} Thanks to their reconstruction objective, auto-encoders compress information in their latent space.
Auto-encoders tend to encode only aspects of the environment that are salient in the input image. 
This means they are not task-specific: relevant elements can be ignored and distractors (unnecessary information) can be encoded into the state representation. They usually need more dimensions that apparently required to encode a scene (e.g. in our experiments, it requires more than 10 dimensions to encode correctly a 2D position).

\item \textit{Reward prediction:} The objective of a reward prediction module leads to state representations that are specialized in a task. However, this does not constrain the state space to be disentangled or to have any particular structure.
\end{itemize}


Combining objectives makes it possible to share the strengths of each model.
In our application example, the previous sections suggest that we should mix objectives to encode both robot and target positions.
The simplest way to combine objectives is to minimize a weighted sum of the different loss functions, i.e. reconstruction, inverse dynamics and reward prediction losses, i.e.: 
$$
\mathcal{L}_{combination}=  w_{reconstruction} \cdot \mathcal{L}_{reconstruction} + w_{inverse} \cdot \mathcal{L}_{inverse} + w_{reward} \cdot \mathcal{L}_{reward}
$$

Each weight represents the relative importance we give to the different objectives. Because we consider each objective to be relevant, we chose the weights such that they provide gradients with similar magnitudes.


\input{ICLR_files/Graphs/full_model.tex}

Combining objectives into a single embedding is not the only option to have features that are \textit{sufficient} to solve the tasks. Stacking representations, which also favors \textit{disentanglement}, is another way of solving the problem. We use this idea in the \textit{SRL Splits} model, where the state representation is split into several parts where each optimizes a fraction of the objectives. This prevents objectives that can be opposed from cancelling out and allows a more stable optimization. This process is similar to training several models but with a shared feature extractor, that projects the observations into the state representation.

In practice, as showed in Fig.~\ref{fig:split-model}, each loss is only applied to part of the state representation. In the experiments, to encode both target and robot positions, we combine the strength of auto-encoders, reward and inverse losses using a state representation of dimension 200. The reconstruction loss is the mean squared error between original and reconstructed image. We used cross entropy loss for the reward prediction and inverse dynamics losses.
The reconstruction and reward losses\footnote{We combine an auto-encoder loss with a reward prediction loss to have task-specific features} are applied on a first split of 198 dimensions and the inverse dynamics loss on the 2 remaining dimensions (encoding the robot position).
To have the same magnitude for each loss, we set $w_{reconstruction}=1$, $w_{reward}=1$ and $w_{inverse}=2$.

The choice of the different hyper-parameters (losses, weights, state dimension, training-set-size) and the robustness to changes are explored and validated in the experiments section (Section ~\ref{sec:experiments}) and Appendix~\ref{sec:additional-res}.

%% file: ICLR_files/Graphs/full_model.tex
\usetikzlibrary{arrows}
\usetikzlibrary{decorations.markings}
\newcommand{\mygrid}{\tikz{\draw[step=0.5cm] (0,0)  grid (0.5,1.5);}}

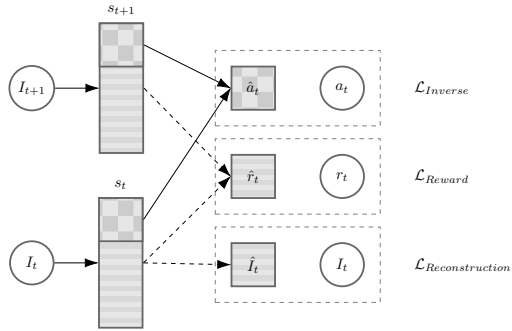
\begin{figure}[ht!]
\centering

\resizebox{0.45\textwidth}{!}{
\begin{tikzpicture}[
roundnode/.style={circle, draw=black!60, fill=green!0, very thick, minimum size=10mm},
roundnode2/.style={circle, draw=black!60, fill=black!20, very thick, minimum size=10mm},
squarednode_st/.style={rectangle, draw=black!60, fill=green!20, very thick, minimum size=10mm},
squarednode/.style={rectangle, draw=black!60, fill=black!20, very thick, minimum size=10mm},
squarednode_st2/.style={rectangle, draw=black!60, very thick, minimum width=10mm, minimum height = 3cm},
invisible/.style={rectangle , draw=black!0, fill=green!0, very thick, minimum size=10mm},
squarednode_img/.style={rectangle, draw=black!60, fill=black!20, very thick, minimum size=10mm},
grid_node/.style={minimum size=.5cm-\pgflinewidth, outer sep=0pt},
container_AE/.style={draw, rectangle, draw=green!60, dashed, inner sep=1em},
container_Inv/.style={draw, rectangle, draw=blue, dashed, inner sep=1em},
container_Rew/.style={draw, rectangle, draw=red!60, dashed, inner sep=1em},
]


\node[invisible]        (base)        {};
\node[invisible]        (base2)        [above=of base] {};

\node[invisible]        (obs)        {};
\node[invisible]        (obs2)        [above=of obs] {};

\node[roundnode]        (hidden_obs2)        [on grid,above=of base2] {$I_{t+1}$};
\node[roundnode]        (hidden_obs)        [on grid,below=of base] {$I_t$};
\node[invisible]        (hidden_obs3)        [on grid,above=of base] {};



\node[invisible]        (hidden_state)        [right=of hidden_obs3,draw] {};
\node[invisible]        (hidden_state)        [on grid,right=of hidden_state,draw] {};

\node[squarednode_st2] (anode) [right=of hidden_obs,draw, pattern=horizontal lines light gray]{};
\node[squarednode_st] (state) [on grid,above=of anode,draw, pattern=checkerboard light gray]{};
\node[] (label) [on grid, above=0.8cm of state]{$s_{t}$};

\node[squarednode_st2] (anode2) [right=of hidden_obs2,draw, pattern=horizontal lines light gray]{};
\node[squarednode_st] (state2) [on grid,above=of anode2,draw, pattern=checkerboard light gray]{};
\node[] (label2) [on grid, above=0.8cm of state2]{$s_{t+1}$};

\node[squarednode]        (recon_rew)        [right=of hidden_state,draw, pattern=horizontal lines light gray] {$\hat{r}_t$};
\node[squarednode]        (recon_img)        [below=of recon_rew,draw, pattern=horizontal lines light gray] {$\hat{I}_t$};
\node[squarednode]        (recon_act)        [above=of recon_rew,draw, pattern=checkerboard light gray] {$\hat{a}_t$};

\node[roundnode]        (img)        [right=of recon_img,draw] {$I_t$};
\node[roundnode]        (rew)        [right=of recon_rew,draw] {$r_t$};
\node[roundnode]        (act)        [right=of recon_act,draw] {$a_t$};

\node[invisible]        (L_img)        [right=of img,draw] {$\mathcal{L}_{Reconstruction}$};
\node[invisible]        (L_rew)        [right=of rew,draw] {$\mathcal{L}_{Reward}$};
\node[invisible]        (L_act)        [right=of act,draw] {$\mathcal{L}_{Inverse}$};


\draw[-{Latex[length=3mm,width=2mm]}] (hidden_obs.east) -- (anode.west);
\draw[-{Latex[length=3mm,width=2mm]}] (hidden_obs2.east) -- (anode2.west);

\draw[-{Latex[length=3mm,width=2mm]}, dashed] (anode.east) -- (recon_rew.west);
\draw[-{Latex[length=3mm,width=2mm]}, dashed] (anode2.east) -- (recon_rew.west);

\draw[-{Latex[length=3mm,width=2mm]}] (state.east) -- (recon_act.west);
\draw[-{Latex[length=3mm,width=2mm]}] (state2.east) -- (recon_act.west);

\draw[-{Latex[length=3mm,width=2mm]}, dashed] (anode.east) -- (recon_img.west);

\node[container_Inv, fit=(recon_act) (act)] (fwd) {};
\node[container_Inv, fit=(recon_img) (img)] (ae) {};
\node[container_Inv, fit=(recon_rew) (rew)] (ae) {};

\end{tikzpicture}
}

\caption{\textit{SRL Splits} model: combines a reconstruction of an image $I$, a reward ($r$) prediction and an inverse dynamic models losses, using two splits of the state representation $s$. Arrows represent model learning and inference, dashed frames represent losses computation, rectangles are state representations, circles are real observed data, and squares are model predictions.}

\label{fig:split-model}
\end{figure}

%% file: sections/experiments.tex
\section{Experiments and Results}
\label{sec:experiments}

\subfile{sections/environments}

We evaluate the two proposed combination methods:
\textbf{(1) SRL Combination}: The combination of reconstruction, reward and inverse losses is done by averaging them on a single embedding. 
\textbf{(2) SRL Splits}: The model described 
 that combines reconstruction, reward and inverse losses using splits of the state representation.
We compare them with several baselines: end-to-end learning from raw pixels, using the Ground Truth (GT) states (i.e., robot and target positions), supervised learning (state representation trained with GT states), Random features, and an auto-encoder. 






Each state representation has a dimension of 200 and is learned using 20 000 samples collected with a random policy.
The implementation and additional training details can be found in Appendix~\ref{sec:appendix:implem}.

\begin{table}[ht!]
 \centering 
\begin{tabular}{ l|llll}\hline 
\textbf{Environments} & \textit{Nav. 1D Target} & \textit{Nav. 2D Target} & \textit{Arm Random Target}  & \textit{Pseudo-Real Omnibot}\\\hline
Ground Truth  & 211.6 $\pm$ 14.0 & 234.4 $\pm$ 1.3 & 4.2 $\pm$ 0.5 & 243.7 $\pm$ 1.2 \\
Supervised & 189.7 $\pm$ 14.8 & 213.5 $\pm$ 6.0 & 3.1 $\pm$ 0.3 & 243.9 $\pm$ 1.8  \\
\hline
Raw Pixels & 215.7 $\pm$ 9.6 & 231.5 $\pm$ 3.1 & 2.6 $\pm$ 0.3 & 185.2 $\pm$ 7.83 \\
Random Features & 211.9 $\pm$ 10.0 & 208 $\pm$ 6.1 & 4.1 $\pm$ 0.3 & 201.5 $\pm$ 5.7 \\
Auto-Encoder & 188.8 $\pm$ 13.5 & 192.6 $\pm$ 8.9 & 3.4 $\pm$ 0.3 & 230.27 $\pm$ 3.2 \\
\hline
SRL Combination & 216.3 $\pm$ 10.0 & 183.6 $\pm$ 9.6 & 2.9 $\pm$ 0.3 & 216.8 $\pm$ 5.6  \\
SRL Splits & 205.1 $\pm$ 11.7 & 232.1 $\pm$ 2.2 & 3.7 $\pm$ 0.3 &  237.8 $\pm$ 2.1  \\
\hline 
\end{tabular}
\caption{End-to-end vs State Representation Learning: Mean reward performance and standard error in RL (using PPO) per episode (average on 100 episodes) at the end of training for all the environments tested.}
\label{tab:rl-perf-all-envs}
\end{table}

Table~\ref{tab:rl-perf-all-envs} displays the mean reward, averaged on 100 episodes, for each environment after RL training. 




\begin{table} [ht!]
 \centering 
\begin{tabular}{ l|llll|l|l}\hline 
\textbf{Ground Truth Correlation} & \textit{$x_{robot}$} & \textit{$y_{robot}$} & \textit{$x_{target}$}  & \textit{$y_{target}$} & \textit{Mean} & \textit{Mean Reward} \\\hline
Ground Truth  &  1 &  1 & 1 & 1 & 1 & 243.7 $\pm$ 1.2 \\
Supervised & 0.69 & 0.73 & 0.6 & 0.61 & 0.66 & 243.9 $\pm$ 1.8 \\
\hline
Random Features & 0.59 & 0.54 & 0.50 & 0.42 & 0.51 & 201.5 $\pm$ 5.7 \\
Robotic Priors & 0.1 & 0.1 & 0.45 & 0.54 & 0.30 & -1.1 $\pm$ 2.4 \\
Auto-Encoder & 0.50 & 0.54 & 0.20 & 0.25  & 0.37 & 230.27 $\pm$ 3.2 \\
\hline
SRL Combination & 0.95 & 0.96 & 0.22 & 0.20 & 0.58 & 216.8 $\pm$ 5.6 \\
SRL Splits & 0.98 & 0.98 & 0.61 & 0.73 & 0.83 & 237.8 $\pm$ 2.1 \\
\hline 
\end{tabular}
\caption{\gtcorr, \gtcorrmean, and mean reward performance in RL (using PPO) per episode after 5 millions steps, with standard error (SE) for each SRL method in 2D Simulated omnibot with a random target environment.}
\label{tab:gt-correlation-omnirobot-2d}
\end{table}


\textit{SRL Splits} is the approach that performs on par or better than learning from \text{raw pixels} across all the tasks. Its counterpart, \textit{SRL Combination}, that uses only a single embedding, gives also positive results, except for the navigation environment with a 2D random target where it under-performs. The correlation with ground truth (\gtcorr in Table~\ref{tab:gt-correlation-omnirobot-2d}, see definition in Appendix) provides us with some insights (see Tables~\ref{tab:gt-correlation-omnirobot-2d}, ~\ref{tab:gt-correlation-mobile-2d}): both methods extract the robot position (absolute correlation close to 1), yet the target position is better encoded with the SRL splits method, which may explain the gap in performance.

In the robotic arm setting, combining approaches does not seem to be of much benefit. Two possible reasons may explains that. First, compared to the mobile robot, the robotic arm and the target are visually salient so an auto-encoder is sufficient to solve the task. Second, the actions magnitude is much smaller in the robotic arm environments, therefore learning an inverse model is much harder in this setting.

\begin{figure}[ht!]
 \centering 
 \includegraphics[width=13.5cm]{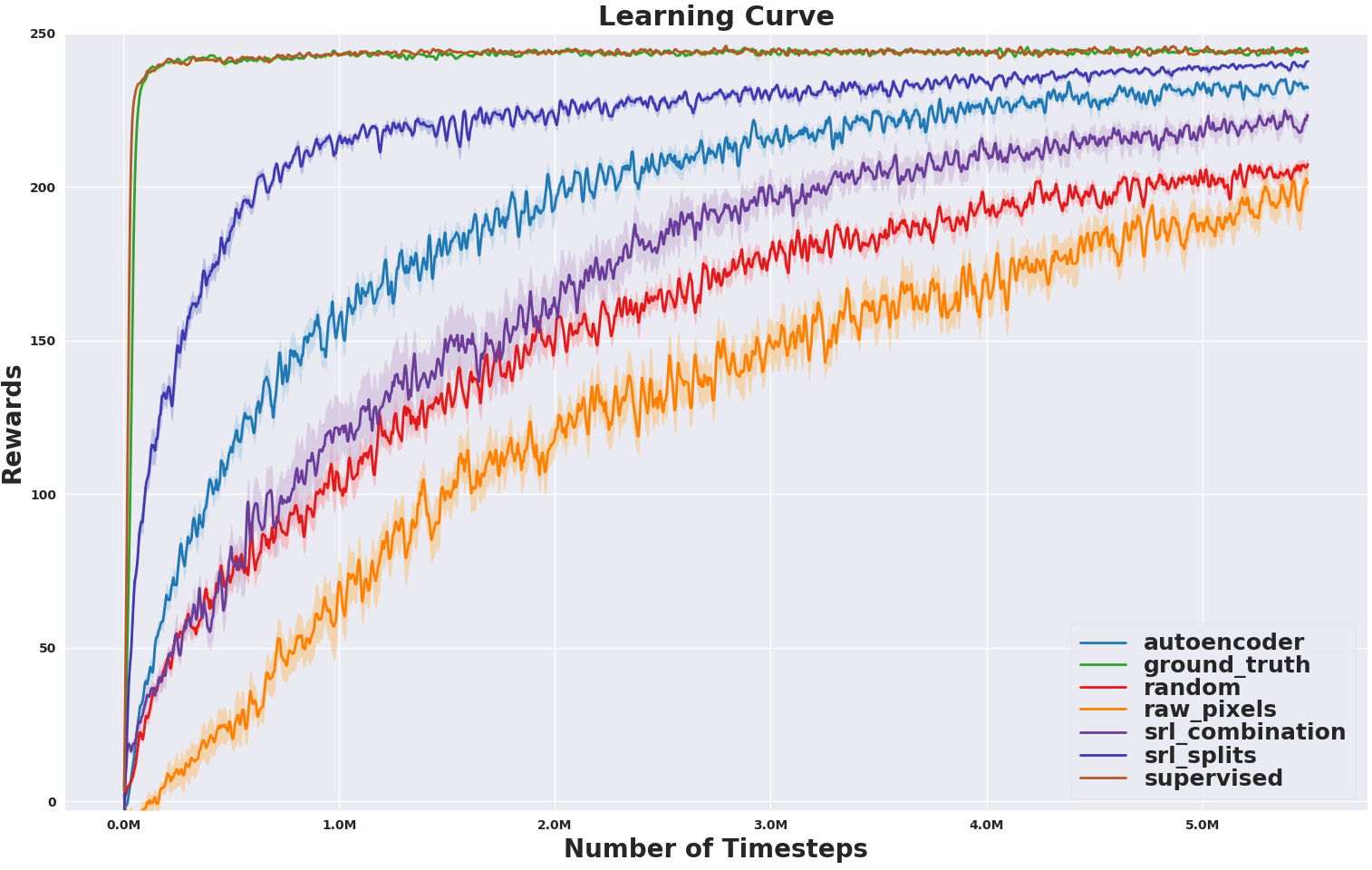} 
 \caption{Performance (mean and standard error for 8 runs) for PPO algorithm for different state representations learned in Simulated OmniRobot with randomly initialized target environment.}
 \label{fig:ppo2-pseudo-omnirobot-random-target}
\end{figure}

\textit{Ground Truth} states naturally outperform all the methods across all environments. This highlights the importance of having a low dimensional and informative representation. 
The \textit{Supervised} baseline allows to quickly attain an acceptable performance, but then reaches a plateau (e.g. Figs.~\ref{fig:ppo2-pseudo-omnirobot-random-target}, ~\ref{fig:ppo2-mobile-2d-target}). Compared to the unsupervised methods, it apparently generalizes less efficiently to data not present in the training set.

As in \cite{Burda2018}, the \textit{Random Features} model performs decently on all the environments and sometimes better 
than learned features. Looking at the \gtcorr (Tables
\ref{tab:gt-correlation-mobile-2d},
\ref{tab:gt-correlation-mobile-1d},  \ref{tab:gt-correlation-kuka-target}), 
random features keep the useful information to solve the tasks. Hence, we hypothesize that random features should work in environments where visual observations are simple enough because they can preserve enough information.

Despite good results in mobile robot navigation with a static target~\citep{Jonschkowski15}, \textit{Robotic Priors} are not well suited when the target changes from episode to episode as they fail to encode the target position. As described in~\cite{Lesort19}, robotics priors lead to a state representation that contains one cluster per episode, which prevent generalization and good performances in these RL tasks.

The \textit{auto-encoder} has mixed results: it allows to solve all environments, yet it sometimes under-performs in the navigation tasks (Fig.~\ref{fig:ppo2-mobile-2d-target}). When we explored the latent space using the S-RL Toolbox visualization tools, we noticed that one dimension of the state space could act on both robot and target positions in the reconstructed image.  Our hypothesis, also supported by the \gtcorr, is that the state space is not disentangled. This approach does not make use of additional information that the environment provides, such as actions and rewards, leading to a latent space that may lack of informative structure.

In the appendix, extra results (Figs.~\ref{fig:random-seed}, \ref{fig:influence-state-dim} and \ref{fig:influence-training-set-size}) exhibit the stability and robustness of SRL against additional hyper-parameter changes (random seed, training set size and dimensionality of the state learned). The state dimension needs to be large enough (at least 50 dimensions for the mobile navigation environment), but increasing it further has no incidence on the performance in RL. In a similar way, a minimal number of training samples ($10 000$) is required to efficiently solve the task. Over that limit, adding more samples does not affect the final mean reward.


During our experiments, we found that learning the policy end-to-end was more sensitive to hyper-parameter changes. For instance, hyper-parameters tuning of A2C (Fig.\ref{fig:a2c-mobile-2d-target}) was needed in order to have decent results for the pixels, whereas the performance was stable for the SRL methods. This can be explained by the reduced search space: the task is simpler to solve when features are already extracted. A more in-depth study would be interesting in the future.

Additionally, results of policy replay on the Simulated \& Real Omnibot (Fig. ~\ref{fig:sim-to-real-omnirobot}) suggest that policies having an efficient state representation trained until near convergence in a high fidelity simulator (\textit{Ground Truth, Supervised, auto-encoder, SRL Splits}) have a more stable behaviour in real life than policies based on raw data or lower performing state representations (\textit{SRL combination, random, raw pixels}).


%% file: sections/environments.tex

\begin{figure}[htbp!]
\centering
  \begin{tabular}
      {ccccc} \hline
      Mobile Navigation &
      Robotic Arm & Simulated \& Real Omnibot
      \\  \hline \\[0.0005pt]
 
\includegraphics[height=0.8in]{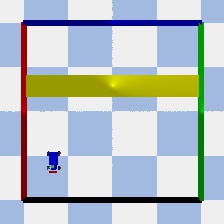} \includegraphics[height=0.8in]{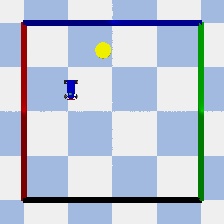} & 
\includegraphics[height=0.8in]{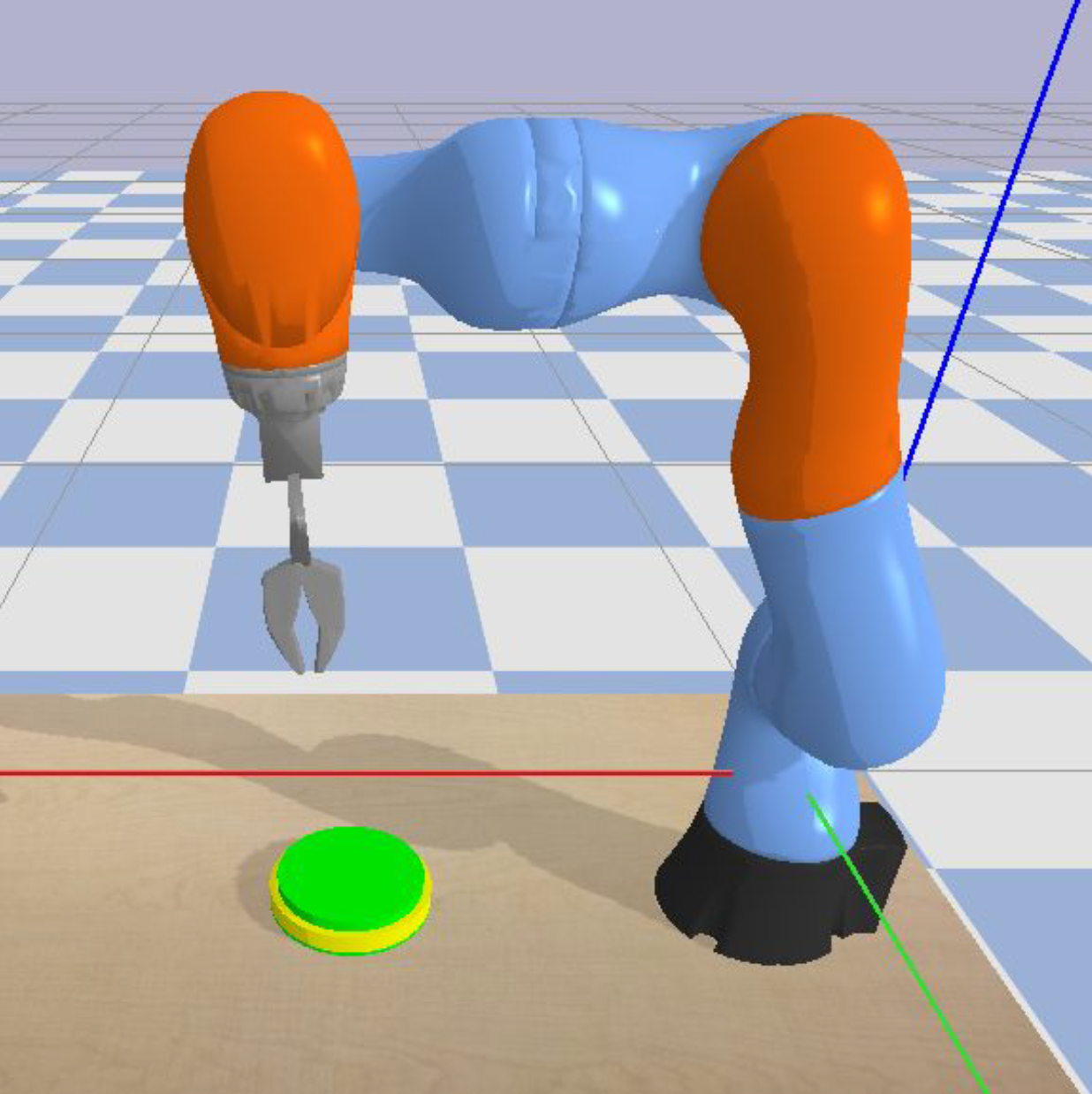} &
\includegraphics[height=0.8in]{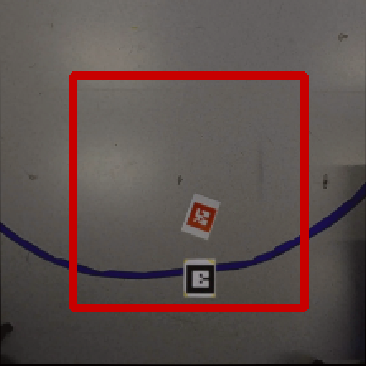}
\includegraphics[height=0.8in]{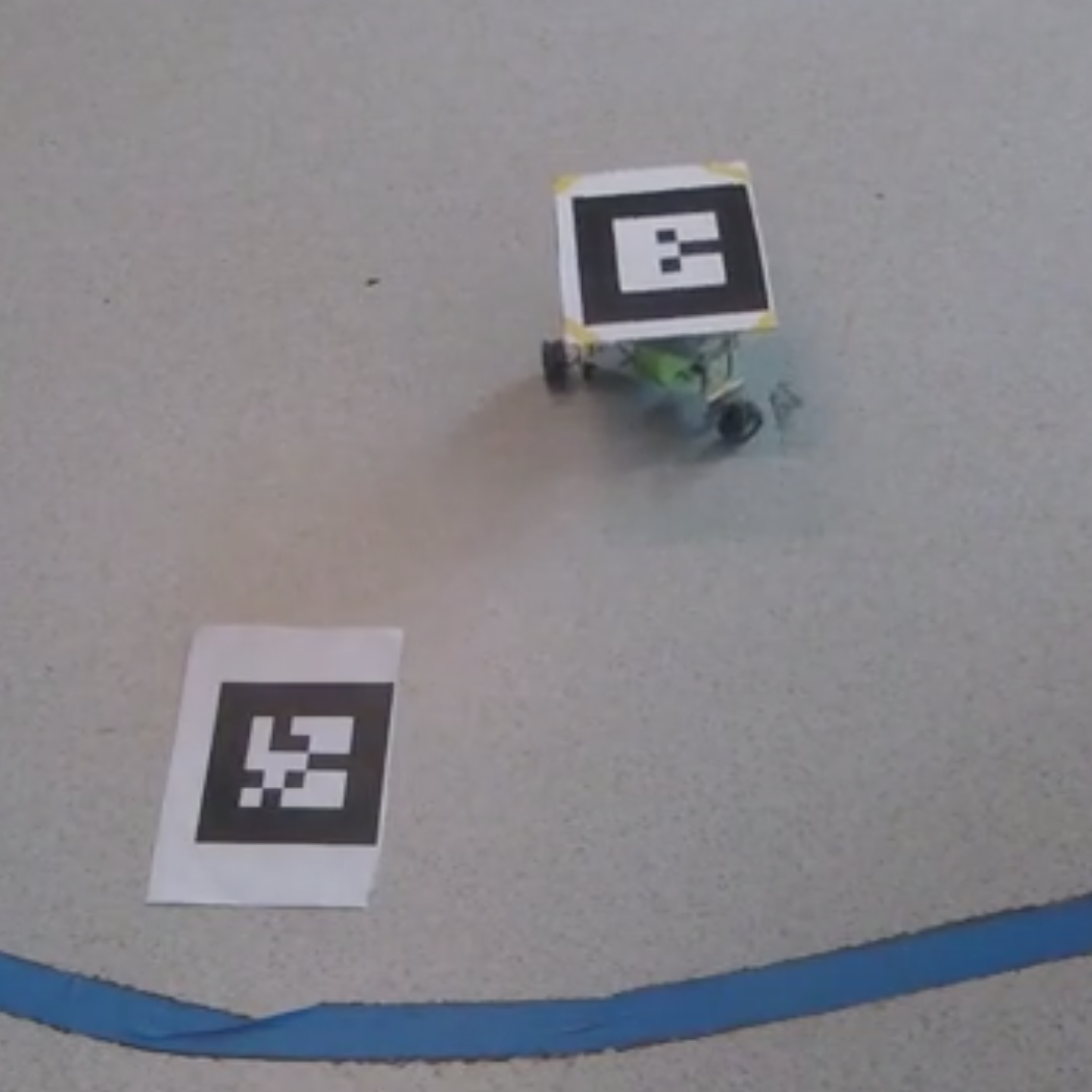}
\\ \hline
  \end{tabular} 
  \caption{Environments for state representation learning from S-RL toolbox~\citep{Raffin18} with extensions (\textit{2D Simulated + Real Omnibot}). 
  }
  \label{fig:datasets}
\end{figure}

In order to evaluate the methods, we use 5 
environments proposed in \textit{S-RL Toolbox}~\citep{Raffin18} (which uses open-source PyBullet \citep{Coumans18}). These can be seen in Fig. \ref{fig:datasets}.
These environments of variable difficulty are specially designed for evaluating SRL methods in a robotics context.
The environments are variations of two main settings: a 2D environment with a mobile robot and a 3D environment with a robotic arm.
In all settings, there is a controlled robot and one target that is randomly initialized. In the experiments, the robot is controlled using discrete actions (but the approaches we present are not limited to that domain) and the reward is sparse: +1 when reaching the goal, -1 when hitting an obstacle and 0 everywhere else. The five environments used are: 1D/2D random target with mobile robot, a random target with robotic arm, a 2D random target with an omnibot in simulation and real life.  Additional details about the environments are available in the section \ref{sec:appendix:environments} of the appendix.

All environments correspond to a fully observable Markov Decision Process (MDP), i.e., target object and agent are always visible and the next observation $o_{t+1}$ only depends on the previous couple ($o_t$, $s_t$)  (except for the robot arm setting with moving target where there is small uncertainty for the position of the target). We chose those specific tasks because they were i) designed for robotics, ii) appropriate for evaluating SRL models iii) of gradual difficulty. Those environments cover basic goal-based robotics tasks: mobile navigation and reaching a 3D position. Then, they are designed with simplicity in mind, making the extracted features easier to interpret. It is also clear what a good state representation should encode because of the small number of relevant elements.

%% file: sections/conclusions.tex
\section{Conclusions}
\label{sec:conclusions}





In this work, we presented the advantages of decoupling feature extraction from policy learning in RL, on a set of goal-based robotics tasks. This decomposition reduces the search space, accelerates training, does not degrade final performances and gives more easily interpretable representations with respect to the true state of the system. We show also that random features provide a good baseline versus end-to-end learning.

We introduced a new way of effectively combining approaches by splitting the state representation. This method uses the strengths of different SRL models and reduces interference between opposed or conflicting objectives when learning a feature extractor.
Finally, we showed the influence of hyper-parameters on SRL Split model, the relative robustness of this model against perturbations, and how SRL models help transfer to a real robot. Future work should confirm these results by experimenting further with real robots in more complex tasks.


%% file: sections/appendix.tex
\newpage
\appendix
\label{sec:appendix}

\section*{Acknowledgments}
This work is supported by the DREAM project\footnote{\url{http://www.robotsthatdream.eu}} through the European Union Horizon 2020 FET research and innovation program under grant agreement No 640891. 

Experiments presented in this paper were carried out using the PlaFRIM experimental testbed, supported by Inria, CNRS (LABRI and IMB), Université de Bordeaux, Bordeaux INP and Conseil Régional d'Aquitaine (see \url{https://www.plafrim.fr/}).

We thank Chuan Qin and Benoit Sarthou for great help with the robotic setup and experiments.

\section{Implementation Details}
\label{sec:implementation}
\label{sec:appendix:implem}
Each state representation is learned using 20 000 samples collected using a random policy. We kept for each method the model with the lowest validation loss during the 30 training epochs. We used the same network architecture from~\citep{Raffin18} for all the models. The input observations of all models are RGB images of size $224 \times 224 \times 3$. Navigation environments use 4 discrete actions (\textit{right, left, forward, backward}); robotic arm environments use one more (\textit{down}) action.

We used PPO (the GPU version called PPO2) and A2C implementations from stable-baselines~\citep{stable-baselines}, a fork of OpenAI Baselines~\citep{Dhariwal17}.
PPO was the RL algorithm that worked well across environments and methods without any hyper-parameter tuning, and therefore, the selected one for our experiments.

Regarding the network learning the policies, the same architecture is used in all different methods. For the approaches that do not use pixels, it is a 2-layers MLP, whereas for learning from raw pixels, it is the CNN from \citep{Mnih15} present in OpenAI baselines.

Observations are normalized, either by dividing the input by 255 (for raw pixels) or by computing a running mean/std average (for SRL models).

For the \textit{SRL Splits} and \textit{SRL Combination} methods, we used a linear model for the inverse dynamics, and a 2-layers MLP of 16 units each with ReLU activation for the reward prediction.
Only one minor adjustment was made on the \textit{SRL Splits} method for the robotic arm environments: because the controlled robot is more complex, the inverse model was allocated more dimensions (10 instead of 2) but keeping the total state dimension constant (equal to 200).

Environments, code and data are available at \url{https://github.com/araffin/robotics-rl-srl}.


\section{Environments}
\label{sec:appendix:environments}

\textit{1D/2D random target mobile navigation}: This environment consists of a navigation task using a mobile robot, similar to the task in \citep{Jonschkowski15}, with either a cylinder (2D target) or a horizontal band (1D target) on the ground as a goal, randomly initialized at the beginning of each episode. The mobile robot can move in four directions (\textit{forward, backward, left, right}) and will get a +1 reward when reaching the target, -1 when hitting walls, and 0 otherwise. Episodes have a maximum length of 250 steps (hence, an upper bound max. reward of 250).

 \textit{Robotic arm with random 
 target}: In this setting, a robotic arm, fixed to a table, has to reach a randomly initialized target on the table. The target can be static during the episode or slowly moving back and forth along one axis. The arm is controlled in the $x$, $y$ and $z$ position using inverse kinematics. The agent receives a +1 reward when it reaches the goal, -1 when hitting the table, and 0 otherwise. The episode terminates either when the robot hits the table or when it touches 5 times the target (hence, the max. reward value is 5). Episodes have a maximum length of 1000/1500 steps in the random/moving target settings, respectively.

\textit{2D random target Simulated and Real Omnibot}:
This environment consists of a navigation task using a 3 wheel omni-directional robot, is similar to the 2D random target mobile navigation 
(identical reward setting and possibility of movement) and available in simulation and real life. 
The robot and the target are identified respectively by black and red QR codes, 
the scene is recorded from above.

\section{Evaluation metrics}

We use two methods to evaluate a learned state representation. First, since the main goal of extracting relevant features is to solve a task, we compare performance in Reinforcement Learning. To have quantitative results, each RL experiment uses 10 different random seeds\footnote{Except the ablation study that uses 5 random seeds}. We chose two metrics: mean reward over 100 episodes at the end of training and mean reward over 100 episodes for a given budget (a fixed number of timesteps). This last metric is particularly relevant when doing robotics: the budget is much more limited than in simulation and we want to reach an acceptable performance as soon as possible.

Then, since we have access to the true positions, we can also compute the correlation between ground truth states and learned states. However, looking at a correlation matrix when the state dimension is large is impractical.
Therefore, we use the metric \textit{GroundTruthCorrelation} (\gtcorr) described in~\citep{Raffin18}.
It measures the maximum correlation (in absolute value) in the learned representation for each dimension of the ground truth (GT) states. 
\gtcorr$\ $ is defined as:
\begin{equation}
GTC_{(i)} = \max\limits_j |\rho_{s, \tilde{s}} (i,j)| \in [0,1] 
\label{eq:c-vect}
\end{equation}

where $\rho_{s, \tilde{s}}$ is the Pearson correlation coefficient for the pair ($s$, $\tilde{s}$), where $\tilde{s}$ is the GT state, $s$ the learned state, $i \in \llbracket0, |\tilde{s}|\rrbracket$, $j \in \llbracket0,|s|\rrbracket$,
$\tilde{s} = [\tilde{s}_{1};...; \tilde{s}_{n}]$, and $\tilde{s}_{k}$ being the $k^{th}$ dimension of the GT state vector. 

\gtcorr produces a vector of a fixed size for a given environment making it able to compare different approaches and different representation dimensions. 

This measure allows to have some understanding of what was learned. More precisely, \gtcorr can provide us with information on the sufficiency: it gives insights of which component of the Ground Truth state was encoded in the state representation. If in addition, the representation is disentangled, then each factor of variation will be encoded by a different component, leading to a \gtcorr close to 1.

However, this measure is only a proxy to minimally guarantee that the required information is encoded: having a good \gtcorr is sufficient to succeed in RL but not necessary. 
It is important to note that this metric is not an estimate of the representation compactness since it is independent of the representation dimension.


\section{Additional Results}
\label{sec:additional-res}
\subsection{Mobile Navigation with 1D Target}

\begin{figure}[ht!]
 \centering 
 \includegraphics[width=13cm]{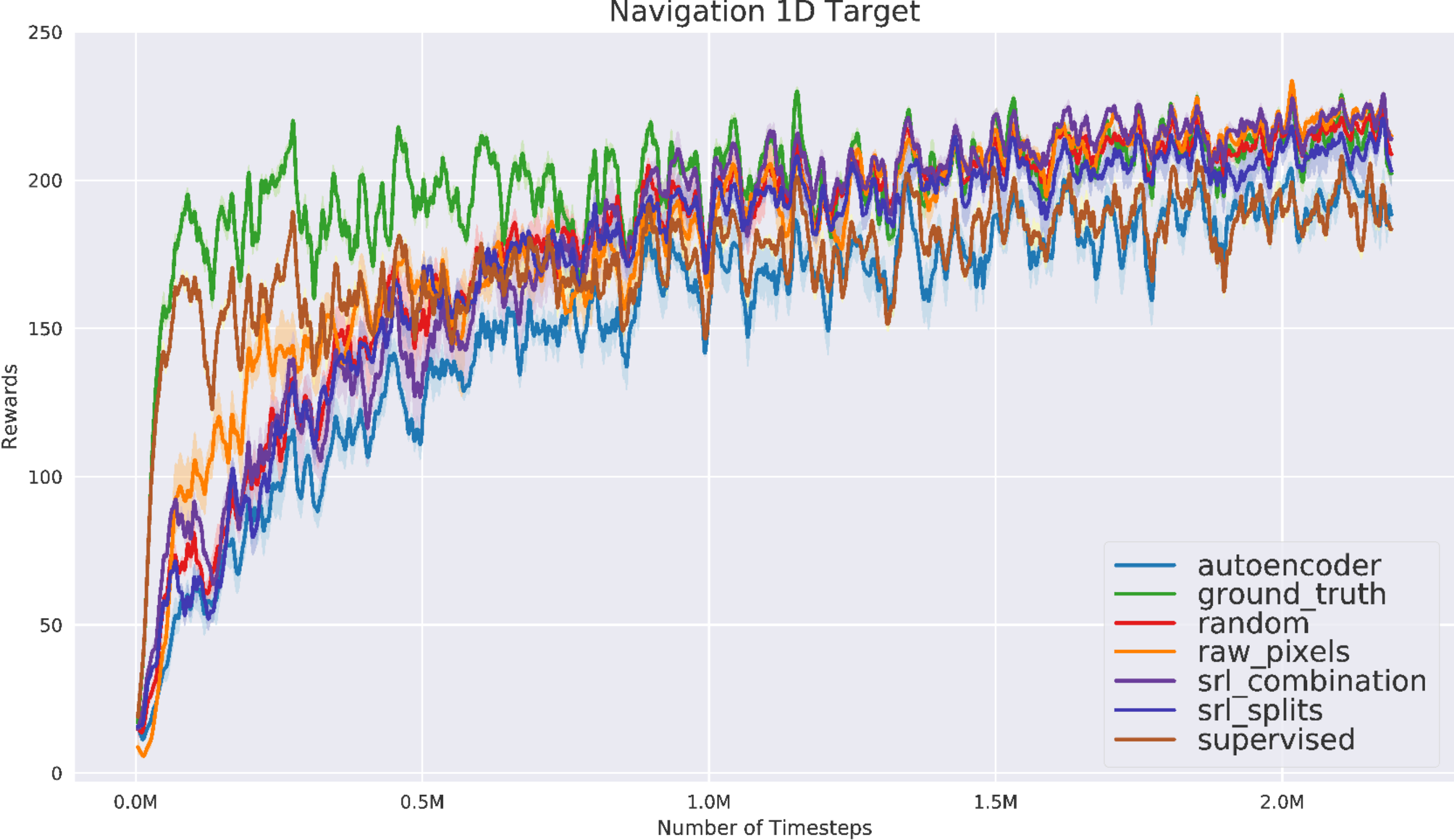}
 \caption{Performance (mean and standard error for 10 runs) for PPO algorithm for different state representations learned in Navigation 1D target environment.}
 \label{fig:ppo2-mobile-line-target}
\end{figure}

\begin{table} [ht!]
 \centering 
\begin{tabular}{ l|lll|l|l}\hline 
\textbf{Ground Truth Correlation} & \textit{$x_{robot}$} & \textit{$y_{robot}$} & \textit{$x_{target}$} & \textit{Mean} & \textit{Mean Reward} \\\hline
Ground Truth  &  1 &  1 & 1 & 1 &  211.6 $\pm$ 14.0 \\
Supervised & 0.68 & 0.77 & 0.72 & 0.72 & 189.7 $\pm$ 14.8 \\
\hline
Random Features & 0.56 & 0.63 & 0.70 & 0.63 & 211.9 $\pm$ 10.0 \\
Auto-Encoder & 0.36 & 0.46 & 0.77 & 0.53 & 188.8 $\pm$ 13.5 \\
\hline
SRL Combination & 0.95 & 0.98 & 0.39 & 0.77 & 216.3 $\pm$ 10.0 \\
SRL Splits & 0.81 & 0.92 & 0.79 & 0.84 & 205.1 $\pm$ 11.7 \\
\hline 
\end{tabular}
\caption{\gtcorr, \gtcorrmean, and mean reward performance in RL (using PPO) per episode after 2 millions steps, with standard error (SE) for each SRL method in Navigation 1D target environment.}
\label{tab:gt-correlation-mobile-1d}
\end{table}

\begin{table} [htbp!]
\centering 
\begin{tabular}{ l|ll}\hline 
\textbf{Budget} (in timesteps) & \textit{1 Million} & \textit{2 Million}\\\hline
Ground Truth  & 198.0 $\pm$ 16.1 & 211.6 $\pm$ 14.0  \\
Supervised & 169.5 $\pm$ 13.5 & 189.7 $\pm$ 14.8 \\
\hline
Raw Pixels & 177.9 $\pm$ 15.6 & 215.7 $\pm$ 9.6 \\
Random Features & 187.8 $\pm$ 12.6 & 211.6 $\pm$ 10.0 \\
Auto-Encoder & 159.8 $\pm$ 16.1 & 188.8 $\pm$ 13.5  \\
\hline
SRL Combination & 191.0 $\pm$ 14.2 & 216.3 $\pm$ 10.0  \\
SRL Splits & 184.5 $\pm$ 12.3 & 205.1 $\pm$ 11.7  \\
\hline 
\end{tabular}
\caption{Mean reward performance in RL (using PPO) per episode (average on 100 episodes) for different budgets, with standard error in Navigation 1D target environment.}
\label{tab:rl-perf-mobile-1d}
\end{table}

\clearpage

\subsection{Mobile Navigation with 2D Target}

\begin{table} [ht!]
 \centering 
\begin{tabular}{ l|llll|l|l}\hline 
\textbf{Ground Truth Correlation} & \textit{$x_{robot}$} & \textit{$y_{robot}$} & \textit{$x_{target}$}  & \textit{$y_{target}$} & \textit{Mean} & \textit{Mean Reward} \\\hline
Ground Truth  &  1 &  1 & 1 & 1 & 1 & 234.4 $\pm$ 1.3 \\
Supervised & 0.69 & 0.73 & 0.70 & 0.72 & 0.71 & 213.5 $\pm$ 6.0 \\
\hline
Random Features & 0.68 & 0.65 & 0.34 & 0.31 & 0.50 & 208 $\pm$ 6.1 \\
Robotic Priors & 0.2 & 0.2 & 0.41 & 0.66 & 0.37 & 6.2 $\pm$ 3.1 \\
Auto-Encoder & 0.52 & 0.51 & 0.24 & 0.23 & 0.38 & 192.6 $\pm$ 8.9 \\
\hline
SRL Combination & 0.92 & 0.92 & 0.33 & 0.42 & 0.65 & 183.6 $\pm$ 9.6 \\
SRL Splits & 0.81 & 0.84 & 0.64 & 0.39 & 0.67 & 232.1 $\pm$ 2.2 \\
\hline 
\end{tabular}
\caption{\gtcorr, \gtcorrmean, and mean reward performance in RL (using PPO) per episode after 5 millions steps, with standard error (SE) for each SRL method in mobile robot navigation 2D random target environment.}
\label{tab:gt-correlation-mobile-2d}
\end{table}


\begin{figure}[ht!]
 \centering 
 \includegraphics[width=15cm]{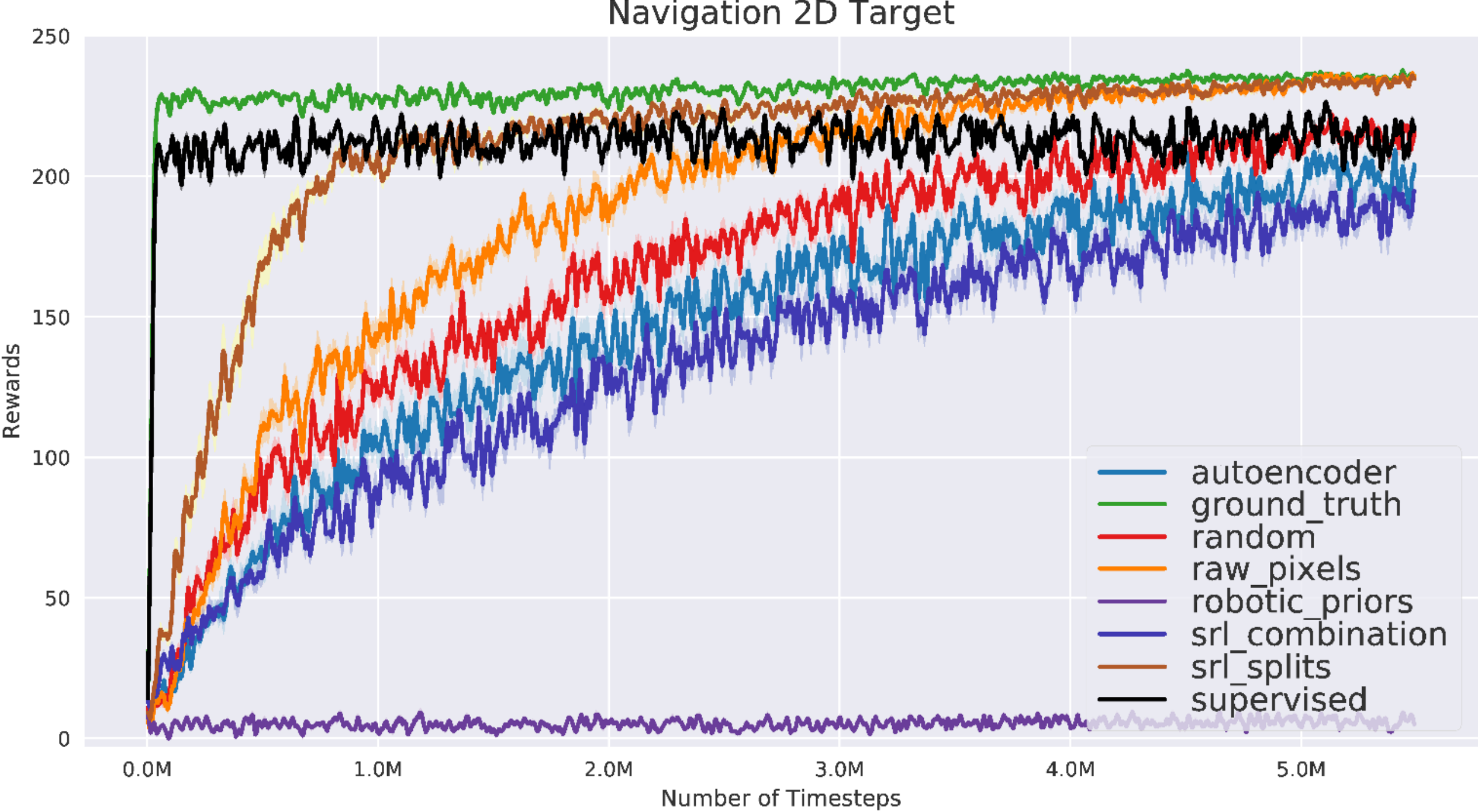}
 \caption{Performance (mean and standard error for 10 runs) for PPO algorithm for different state representations learned in Navigation 2D random target environment.}
 \label{fig:ppo2-mobile-2d-target}
\end{figure}

\begin{figure}[ht!]
 \centering 
 \includegraphics[width=15cm, height=7cm]{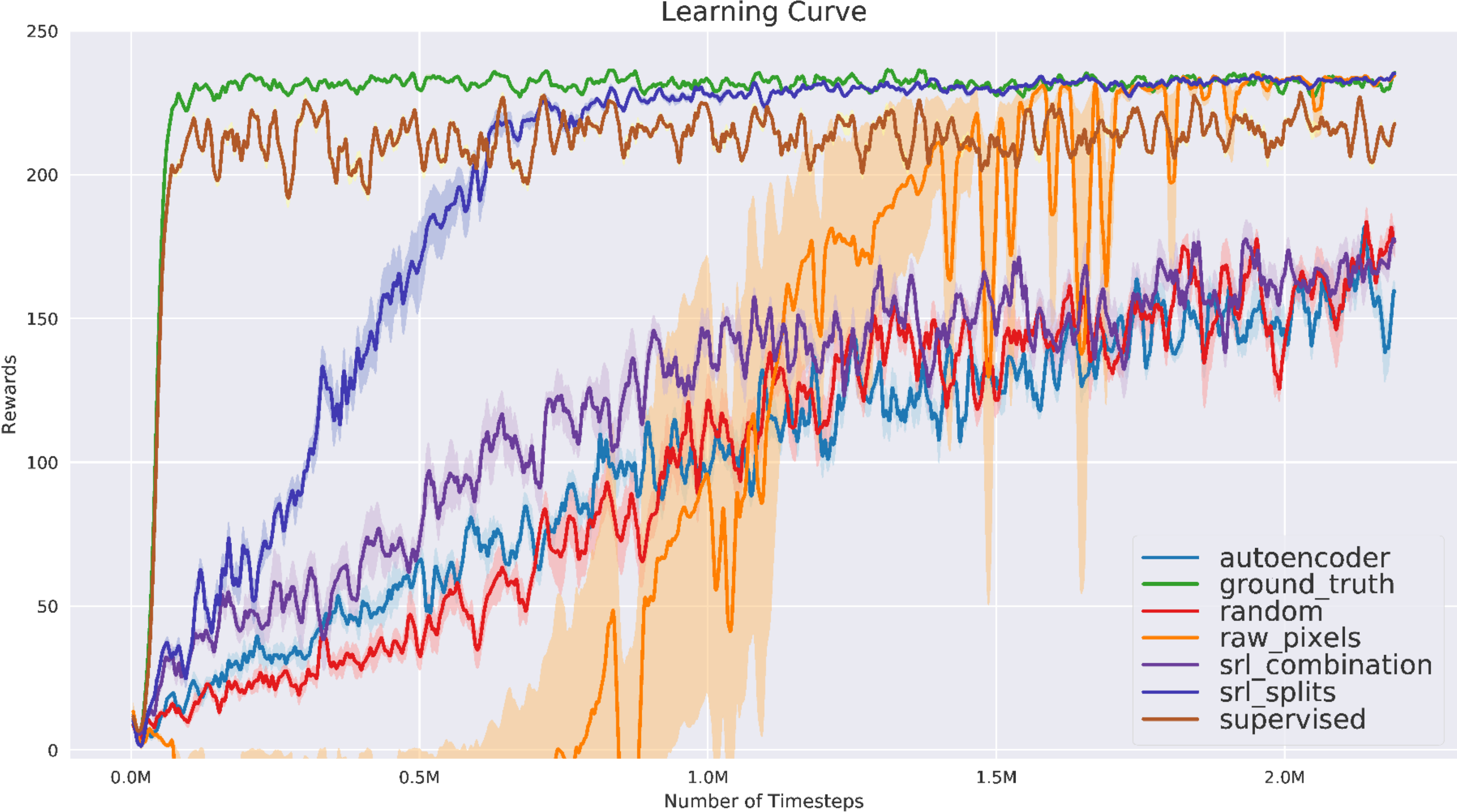}
 \caption{Performance (mean and standard error for 10 runs) for A2C algorithm for different state representations learned in Navigation 2D random target environment.}
 \label{fig:a2c-mobile-2d-target}
\end{figure}

\begin{table} [htbp!]
 \centering 
\begin{tabular}{ l|llll}\hline 
\textbf{Budget} (in timesteps) & \textit{1 Million} & \textit{2 Million} & \textit{3 Million}  & \textit{5 Million}\\\hline
Ground Truth  & 227.8 $\pm$ 2.8 & 229.7 $\pm$ 2.7 & 231.5 $\pm$ 1.9 & 234.4 $\pm$ 1.3 \\
Supervised & 213.1 $\pm$ 6.1 & 213.3 $\pm$ 6.0 & 214.7$\pm$ 5.6 & 213.5 $\pm$ 6.0 \\
\hline
Raw Pixels & 136.3 $\pm$ 11.5 & 188.2 $\pm$ 9.4 & 214.0 $\pm$ 5.9 & 231.5 $\pm$ 3.1 \\
Random Features & 116.3 $\pm$ 11.2 & 163.4 $\pm$ 10.0 & 186.8 $\pm$ 8.2 & 208 $\pm$ 6.1 \\
Robotic Priors & 4.9 $\pm$ 2.9 & 5.4 $\pm$ 3.1 & 4.9 $\pm$ 2.8 & 6.2 $\pm$ 3.1 \\
Auto-Encoder & 97.0 $\pm$ 12.3 & 138.5 $\pm$ 12.3 & 167.7 $\pm$ 11.1 & 192.6 $\pm$ 8.9 \\
\hline
SRL Combination & 83.9 $\pm$ 40.7 & 123.1 $\pm$ 11.6 & 150.1 $\pm$ 11.0 & 183.6 $\pm$ 9.6 \\
SRL Splits & 205.5 $\pm$ 6.6 & 219.5 $\pm$ 5.1 & 223.4 $\pm$ 4.5 & 232.1 $\pm$ 2.2 \\
\hline 
\end{tabular}
\caption{Mean reward performance in RL (using PPO) per episode (average on 100 episodes) for different budgets, with standard error in Navigation 2D random target environment.}
\label{tab:rl-perf-mobile-2d}
\end{table}



\subsection{Robotic arm with randomly initialized target}

\begin{figure}[ht!]
 \centering 
 \includegraphics[width=14cm]{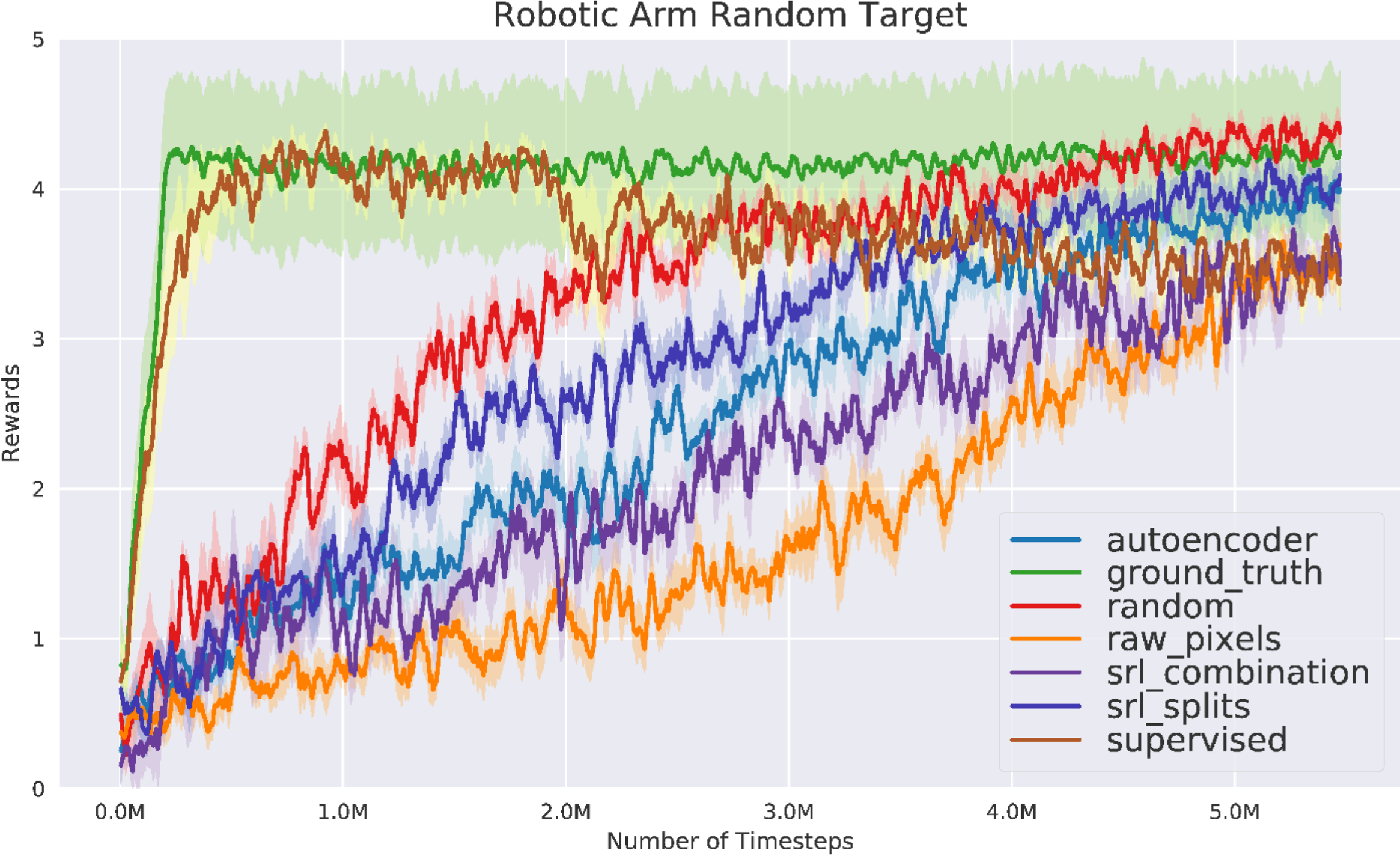}
 \caption{Performance (mean and standard error for 10 runs) for PPO algorithm for different state representations learned in robotic arm with random target environment.}
 \label{fig:ppo2-kuka-target}
\end{figure}

\begin{table} [htbp!]
\centering 
\begin{tabular}{ l|lllll|l|l}\hline 
\textbf{Ground Truth Correlation} & \textit{$x_{robot}$} & \textit{$y_{robot}$} & \textit{$z_{robot}$} & \textit{$x_{target}$}  & \textit{$y_{target}$} & \textit{Mean} & \textit{Mean Reward} \\\hline
Ground Truth  &  1 & 1 & 1 & 1 & 1 & 1 & 4.2 $\pm$ 0.5 \\
Supervised & 0.46 & 0.58 & 1.0 & 0.94 & 0.84 & 0.65  & 3.1 $\pm$ 0.3 \\
\hline
Random Features & 0.34 & 0.58 & 0.62 & 0.71 & 0.83 & 0.55 & 4.1 $\pm$ 0.3 \\
Robotic Priors & 0.21 & 0.18 & 0.42  & 0.7 & 0.66 & 0.38 & 2.6 $\pm$ 0.3 \\
Auto-Encoder & 0.45 & 0.8 & 0.84 & 0.40 & 0.45 & 0.53 & 3.4 $\pm$ 0.3 \\
\hline
SRL Combination & 0.5 & 0.8 & 0.71 & 0.59 & 0.55 & 0.56 & 2.9 $\pm$ 0.3 \\
SRL Splits & 0.42 & 0.81 & 0.73 & 0.51 & 0.58 & 0.55  &  3.7 $\pm$ 0.3 \\
\hline 
\end{tabular}
\caption{\gtcorr, \gtcorrmean, and mean reward performance in RL (using PPO) per episode after 5 millions steps, with standard error for each SRL method in robotic arm with random target environment.}
\label{tab:gt-correlation-kuka-target}
\end{table}


\begin{table}[htbp!]
    \centering
    \begin{tabular}{l|llll}\hline 
    \textbf{Budget} (in timesteps) & \textit{1 Million} & \textit{2 Million} & \textit{3 Million}  & \textit{5 Million}\\\hline
    Ground Truth  & 4.1 $\pm$ 0.5 & 4.1 $\pm$ 0.6 & 4.1 $\pm$ 0.6 & 4.2 $\pm$ 0.5 \\
    Supervised & 4.0 $\pm$ 0.3 & 3.8 $\pm$ 0.3 & 3.4 $\pm$ 0.3 & 3.1 $\pm$ 0.3 \\
    \hline
    Raw Pixels & 0.6 $\pm$ 0.3 & 0.8 $\pm$ 0.3 & 1.2 $\pm$ 0.3 & 2.6 $\pm$ 0.3 \\
    Random Features & 1.5 $\pm$ 0.4 & 2.8 $\pm$ 0.3 & 3.5 $\pm$ 0.3 & 4.1 $\pm$ 0.3 \\
    Auto-Encoder & 0.92 $\pm$ 0.3 & 1.6 $\pm$ 0.3 & 2.2 $\pm$ 0.3 & 3.4 $\pm$ 0.3 \\
    \hline
    SRL Combination & 1.0 $\pm$ 0.3 & 1.5 $\pm$ 0.3 & 2.0 $\pm$ 0.3 & 2.9 $\pm$ 0.3 \\
    SRL Splits & 1.1 $\pm$ 0.3 & 2.1 $\pm$ 0.3 & 2.7 $\pm$ 0.4 & 3.7 $\pm$ 0.3 \\
    \hline 
    \end{tabular}
\caption{Mean reward performance in RL (using PPO) per episode (average on 100 episodes) for different budgets, with standard error in robotic arm with random target environment.}
\label{tab:rl-perf-kuka-target}
\end{table}

Table~\ref{tab:rl-perf-kuka-target} shows the RL performance for different budgets on the Robotic Arm with random target task. To compare SRL methods, \gtcorr, \gtcorrmean and associated RL performance are displayed in Table~\ref{tab:gt-correlation-mobile-2d} for the navigation task with a 2D target. 

\clearpage

\subsection{Ablation study and hyperparameters influence study}
\label{sec:ablation-study}

To better understand the influence of each hyper-parameter and study the robustness of SRL, we performed a thorough analysis of \textit{SRL Splits} in the mobile robot navigation with 2D random target setting.

Figure~\ref{fig:ablation-study} (and Table~\ref{tab:gt-correlation-ablation} in the appendix) show the result of the ablation study performed on the \textit{SRL Splits} model. As expected, the inverse model allows to extract the position of the controllable object, which is the robot. This helps to solve the task and results in a performance boost. In the same vein, the addition of a reward loss favors the encoding of the target position.
It also does not seem necessary to separate the reconstruction and reward losses as they encode the same information.


Table~\ref{tab:influence-weights} displays the influence of the weights of the loss combination on the final mean reward. It shows that the method works on a wide range of different weighting schemes, as long as the reconstruction and the inverse loss have similar magnitude. When the reconstruction weight is one order of magnitude greater, the model behaves like an auto-encoder (because the feature extractor is shared).

\begin{table}[htbp!]
\centering
    \begin{adjustbox}{}
    \centering
    \begin{tabular}{lll|l}
    \hline 
    $w_{reconstruction}$ & $w_{reward}$ & $w_{inverse}$ & \textit{Mean Reward} \\ \hline
    1              & 1      & 1       &  225.2 $\pm$ 6.3 \\
    1              & 1      & 10       & 223.5  $\pm$ 8.0 \\
    1              & 10      & 10       & 215.1  $\pm$ 7.1 \\
    1              & 10      & 5       & 217.8  $\pm$ 12.2 \\
    1              & 5      & 1       & 217.8 $\pm$ 6.7  \\
    1              & 5      & 10       & 228.8 $\pm$  4.2 \\
    5              & 1      & 1       & 221.0 $\pm$ 7.4 \\
    5              & 1      & 10       & 209.1 $\pm$ 19.5 \\
    5              & 10      & 10      & 226.3 $\pm$ 5.2 \\
    5              & 5      & 1       & 194.6 $\pm$ 14.6 \\
    5              & 5      & 10       & 224.5 $\pm$ 5.5 \\
    10              & 1      & 1       & 176.5  $\pm$ 16.2 \\
    10              & 1      & 10       & 218.9 $\pm$ 8.0 \\
    10              & 1      & 5       & 182.4 $\pm$  15.8\\
    10             & 5      & 10       & 225.5 $\pm$ 5.7 \\
    10             & 5      & 5       & 210.2$\pm$ 8.3 \\
    \hline
    \end{tabular}
    \end{adjustbox}
    \caption{Influence of the weights on the SRL Splits model performance, Navigation 2D random target environment.}
    \label{tab:influence-weights}
\end{table}

\begin{table}[htbp!]
\centering 
\begin{tabular}{l|llll|l|l}\hline 
\textbf{Ground Truth Correlation} & \textit{$x_{robot}$} & \textit{$y_{robot}$} & \textit{$x_{target}$}  & \textit{$y_{target}$} & \textit{Mean} & \textit{Mean reward} \\\hline
Auto-Encoder & 0.52 & 0.51 & 0.24 & 0.23 & 0.38 & 138.5 $\pm$ 12.3 \\
Auto-Encoder / Inverse & 0.94 & 0.94 & 0.37 & 0.40 & 0.66 & 185.4 $\pm$ 16.4 \\
Auto-Encoder + Reward & 0.41 & 0.37 & 0.70 & 0.46 & 0.48 & 200.7 $\pm$ 10.1 \\
Reward & 0.57 & 0.43 & 0.32 & 0.57 & 0.47 & 150.1 $\pm$ 15.2 \\
Reward / Inverse & 0.85 & 0.92 & 0.48 & 0.67 & 0.73 & 211 $\pm$ 8.2 \\
Auto-Encoder / Reward / Inverse (SRL 3 Splits) & 0.92 & 0.89 & 0.51 & 0.59 & 0.73 & 223.4 $\pm$ 5.6 \\
Auto-Encoder + Reward / Inverse (SRL Splits) & 0.81 & 0.84 & 0.64 & 0.39 & 0.67 & 232.1 $\pm$ 2.2 \\
Auto-Encoder + Reward / Inverse + Forward & 0.99 & 0.99 & 0.31 & 0.33 & 0.66 & 159.6 $\pm$ 15.1 \\
Raw Pixels & N/A & N/A & N/A & N/A & N/A & 188.2 $\pm$ 9.5 \\
\hline 
\end{tabular}
\caption{\gtcorr, \gtcorrmean, and mean reward performance in RL (using PPO) per episode after 2 millions steps, with standard error for each SRL method in Navigation 2D random target environment. The slash \textit{/} stands for using different splits of the state representation, and the plus \textit{+} for combining methods on a shared representation; e.g \textit{Auto-Encoder + Reward} stands for combining an Auto-Encoder to a Reward model. whereas \textit{Auto-Encoder / Reward} means that each loss applies on a separate part of the state representation.}
\label{tab:gt-correlation-ablation}
\end{table}

\begin{figure}[h]
 \centering 
 \includegraphics[width=12cm]{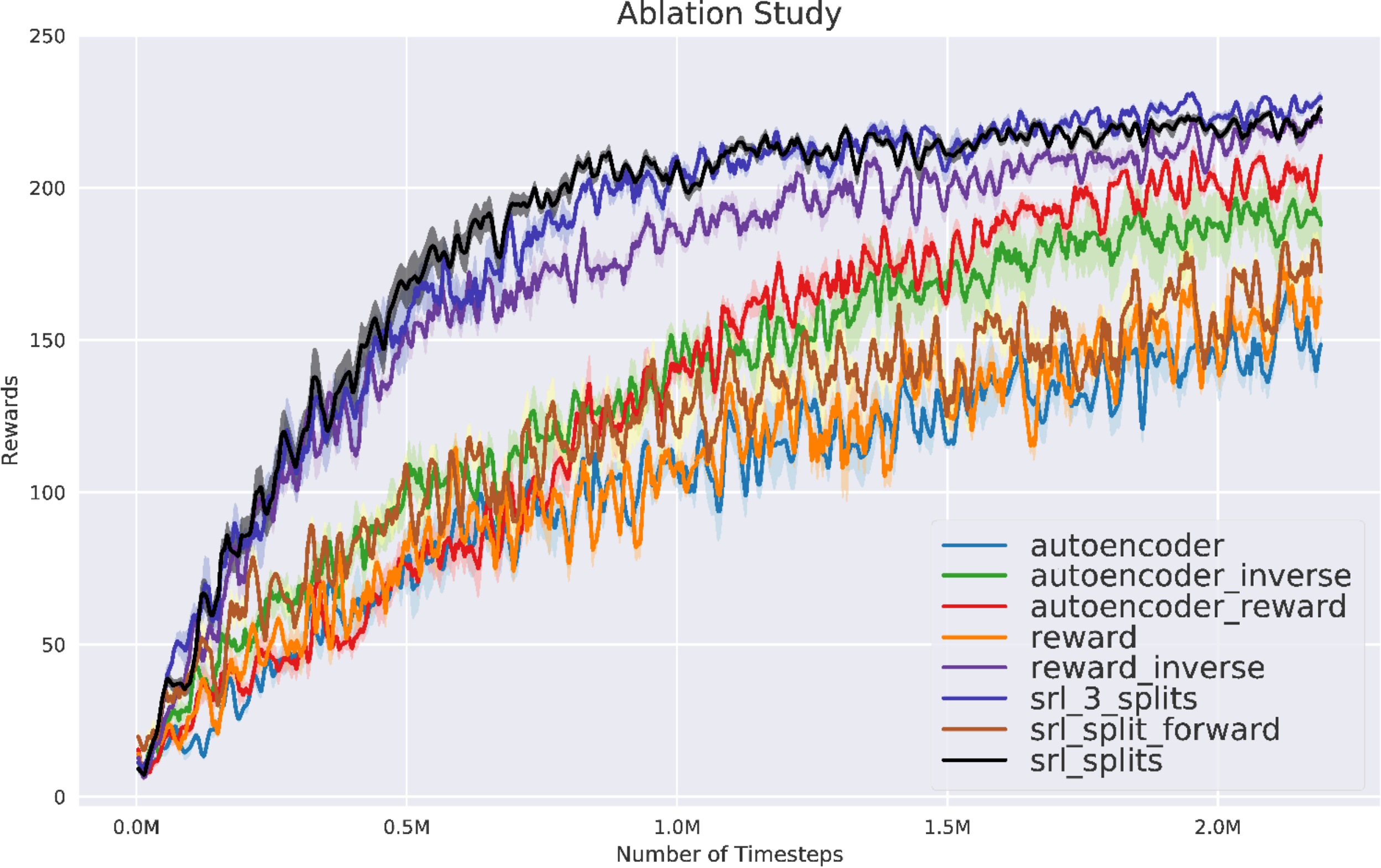}
 \caption{
 Ablation study of \textit{SRL Splits} (mean and standard error for 10 runs) for PPO algorithm in Navigation 2D random target environment. Models details are explained in Table \ref{tab:gt-correlation-ablation},
 e.g.,  \textit{SRL\_3\_splits} model allocates separate parts of the state representation to each loss (reconstruction/reward/inverse).
 }
 \label{fig:ablation-study}
\end{figure}


For every environment, there is always a SRL method that reaches or exceeds the performance obtained using only the raw pixels as input. 
SRL methods do not necessary improve final performance (see Table ~\ref{tab:rl-perf-all-envs}), however as shown in Table~\ref{tab:rl-perf-kuka-target}, it is useful to improve the learning speed of the policy. For instance, in the robotic arm task, much more samples are needed to attain similar levels of performance to those achieved by learning in an end-to-end manner.

\subsection{Influence of the random seed}

\begin{figure}[htbp!]
 \centering 
\includegraphics[width=12cm]{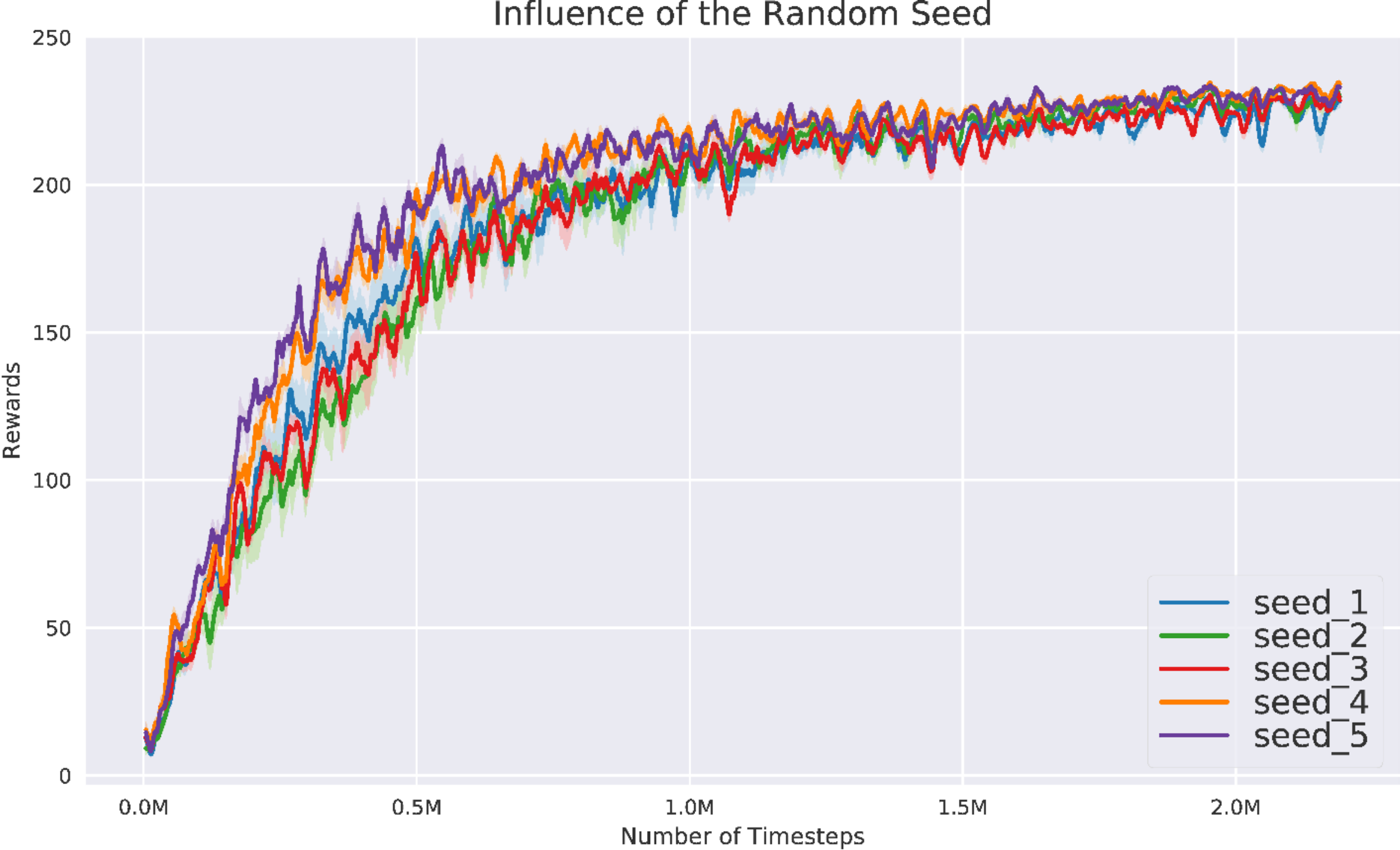}
 \caption{Influence of random seed (mean and standard error for 10 runs) for PPO algorithm for SRL Splits in Navigation 2D random target environment}
 \label{fig:random-seed}
\end{figure}

Figure \ref{fig:random-seed} shows that the SRL \textit{Split} method is stable and its performance does not depend on the random seed used.

\subsection{Influence of the state dimension}

\begin{figure}[htbp!]
 \centering 
 \includegraphics[width=12cm, height=7cm]{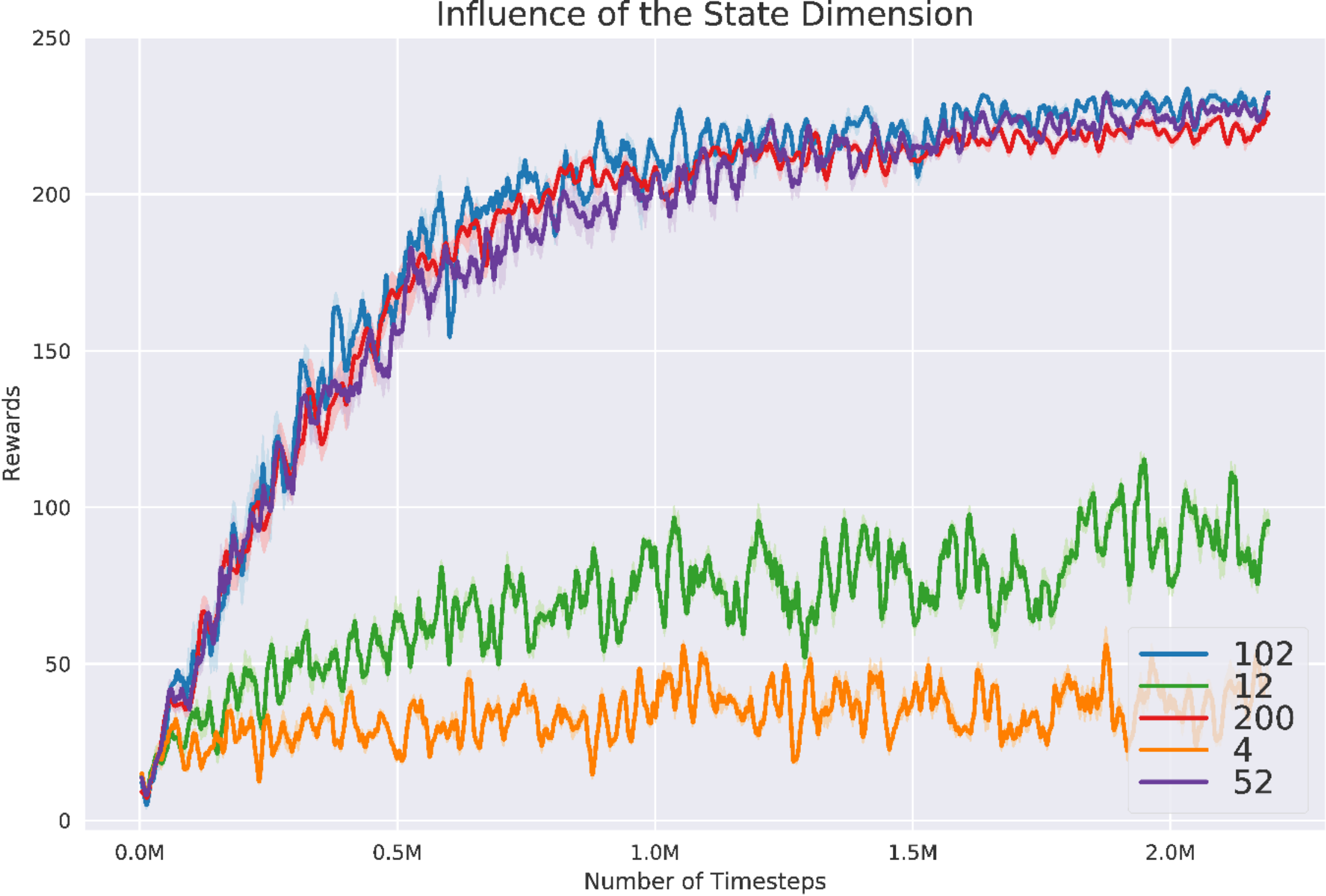}
 \caption{Influence of the state dimension (mean and standard error for 10 runs) for PPO algorithm for SRL Splits in Navigation 2D random target environment. Each label correspond to the state dimension of the model.}
 \label{fig:influence-state-dim}
\end{figure}

As shown in figure~\ref{fig:influence-state-dim}, the state dimension for the SRL model needs to be large enough in order to efficiently solve the task. However, over a threshold, increasing the state dimension does not affect (positively or negatively) the performance in RL.


\subsection{Influence of the training set size}

\begin{figure}[htbp!]
 \centering 
 \includegraphics[width=12cm]{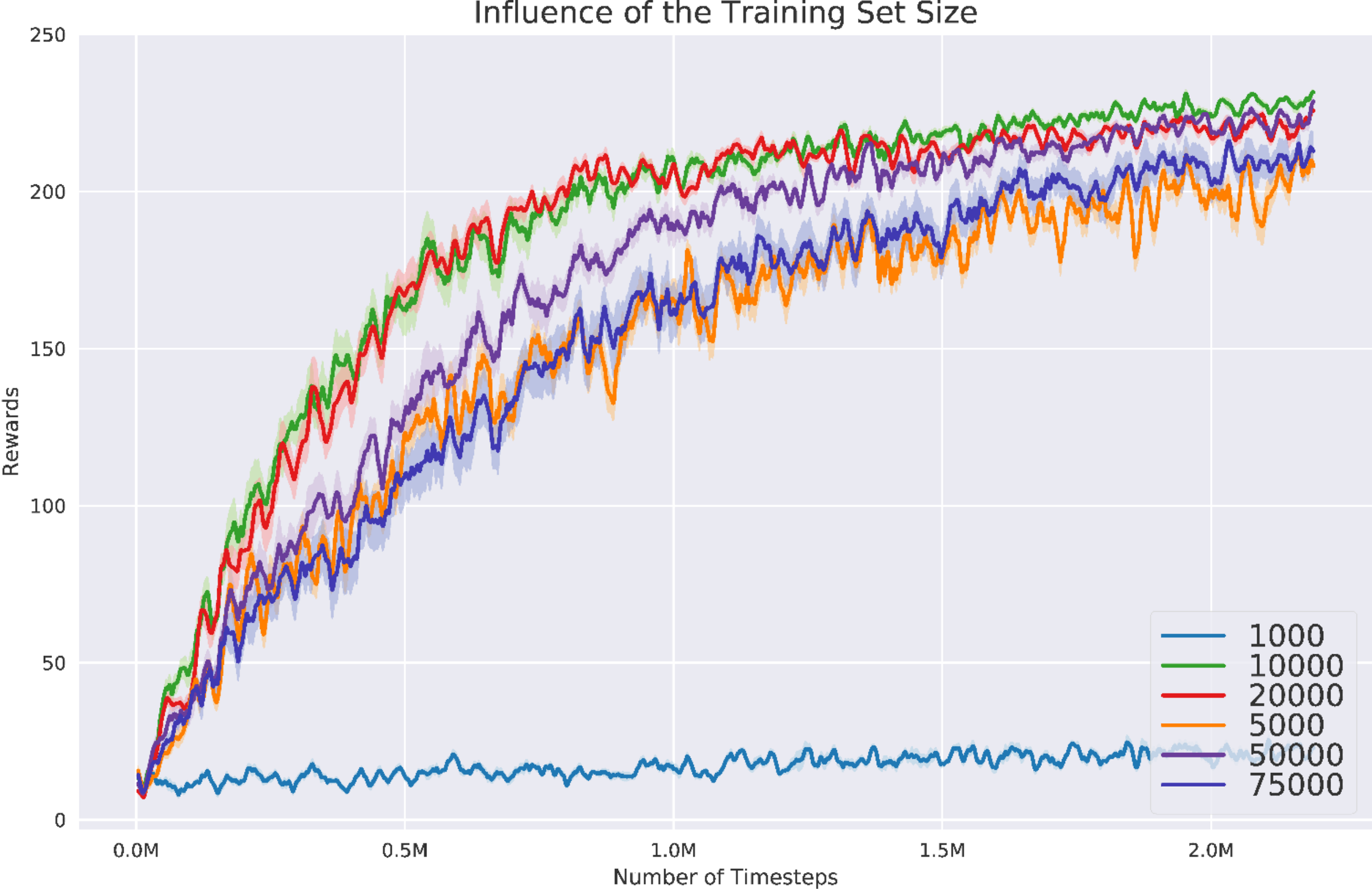}
 \caption{Influence of the training set size (mean and standard error for 10 runs) for PPO algorithm for SRL Splits in Navigation 2D random target environment. Each label corresponds to the number of samples used to train the SRL model.}
 \label{fig:influence-training-set-size}
\end{figure}

The influence of the training set size (Fig.~\ref{fig:influence-training-set-size}) is somehow similar to the influence of the state dimension (Fig.~\ref{fig:influence-state-dim}). A minimal number of training samples is required to solve the task, but over a certain limit, increasing the training set size is not beneficial anymore.
\clearpage

\subsection{From simulation to Real Omnibot: replay of policies.}

\begin{figure}[htbp!]
\centering 
\includegraphics[width=15cm]{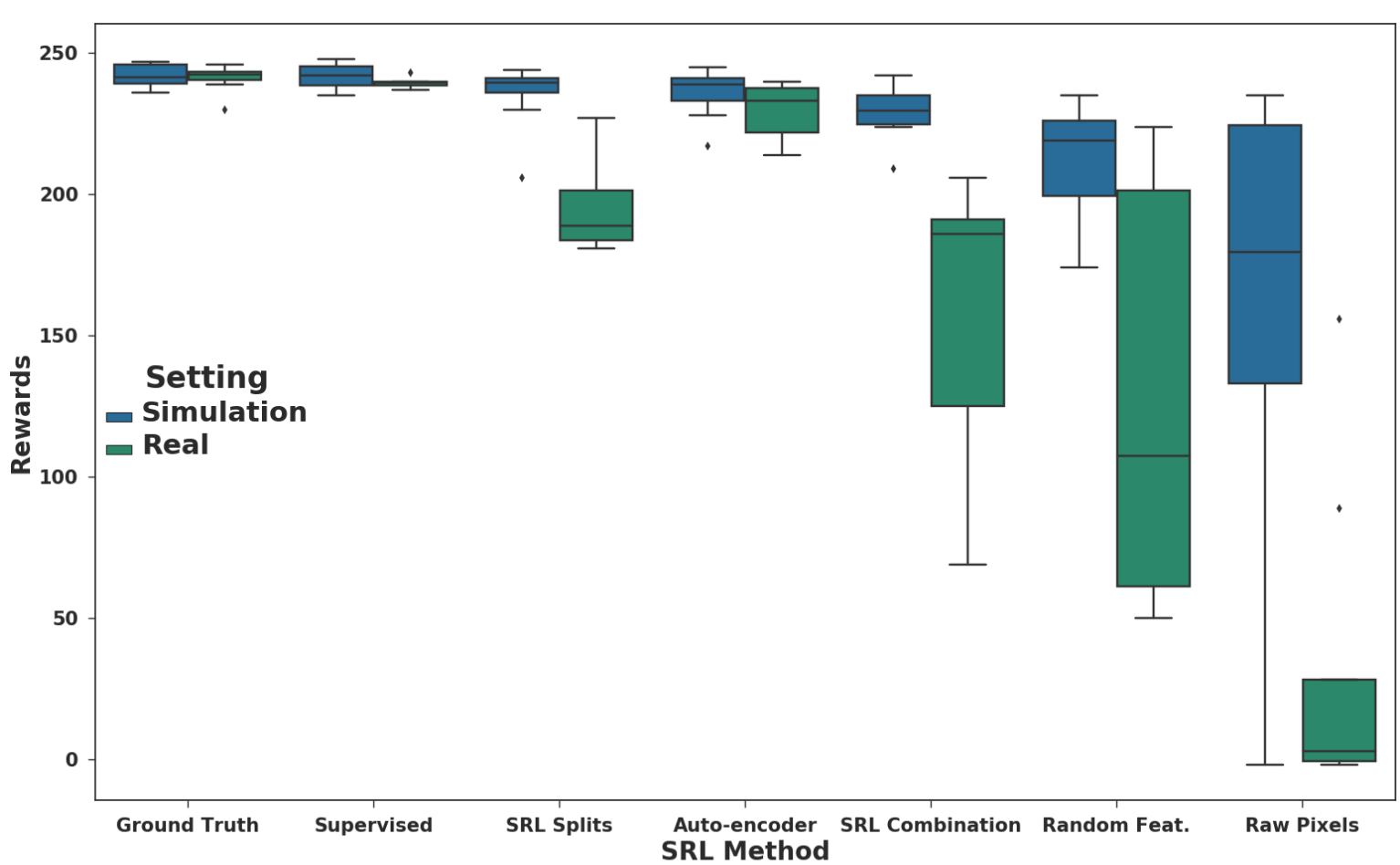} 
\caption{From simulation to real robot: Mean reward and standard deviation for policies trained in simulation (5M steps budget) and replayed in Simulated and Real Omnibot (250 steps, 8 runs).}
\label{fig:sim-to-real-omnirobot}
\end{figure}

Results of policy replay on the Simulated \& Real Omnibot (Fig. ~\ref{fig:sim-to-real-omnirobot}) suggest that policies having an efficient state representation trained until near convergence in a high fidelity simulator (\textit{Ground Truth, Supervised, auto-encoder, SRL Splits}) have a more stable behaviour in real life than policies based on raw data or lower performing state representations (\textit{SRL combination, random features and raw pixels}).